\documentclass{article}
\usepackage{graphicx, geometry, booktabs, amsmath, amsfonts, commath, appendix, authblk}
\title{To impute or not to impute: How machine learning modelers treat missing data}
\author[1]{Wanyi Chen}
\author[2]{Mary Cummings}
\affil[1]{Duke University}
\affil[2]{George Mason University}
\date{}

\begin{document}

\maketitle

\begin{abstract}
    Missing data is prevalent in tabular machine learning (ML) models, and different missing data treatment methods can significantly affect ML model training results. However, little is known about how ML researchers and engineers choose missing data treatment methods and what factors affect their choices. To this end, we conducted a survey of 70 ML researchers and engineers. Our results revealed that most participants were not making informed decisions regarding missing data treatment, which could significantly affect the validity of the ML models trained by these researchers. We advocate for better education on missing data, more standardized missing data reporting, and better missing data analysis tools.
\end{abstract}

\section{Introduction}

Missing data are prevalent in real-world artificial intelligence datasets across many domains. For instance, a study analyzing over 1.7 million cases in the National Cancer Database found that 54.9\% of patients had at least one missing or incomplete variable \cite{yang2020impact}. A survey reviewing 1,666 quantitative studies published in 11 educational journals found that 48\% of the studies involved missing data \cite{peng2006advances}. 

Missing data treatment is an important step in training machine learning (ML) models. Different missing data treatment methods can significantly affect ML model training results and may lead to different conclusions \cite{little2024comparison, shadbahr2023impact}. Moreover, if handled inappropriately, missing data treatment may induce additional biases into ML models \cite{joel2022review, little2024comparison}, as well as call into question their validity. 

However, despite its importance, the treatment of missing data treatment is often either not discussed in papers or only addressed in a curosry fashion \cite{nijman2022missing}. Little is known about how ML modelers choose missing data treatment methods and what factors affect their choices. The lack of awareness and transparency regarding missing data treatment makes it difficult to assess the reproducibility and trustworthiness of ML models where missing data is a factor. 

This problem becomes prominent as ML and artificial intelligence become increasingly popular across various domains. For example, many recognize ML's potential to help scientists discover more scientific knowledge and do so more efficiently \cite{/content/publication/a8d820bd-en}. However, currently ML does not reach this potential, in part because it can be a very subjective process \cite{chen2024subjectivity, cummings2021subjectivity}. To investigate the subjectivity involved in missing data treatment, we conducted a survey of 70 ML researchers and engineers regarding their knowledge and practices of missing data treatment. We aim to shed light on the modelers' preferences in missing data treatment and its implications in responsible model training.

\section{Background and related work}

Missing data occur for different reasons. Statistically, they are classified by three different missing data mechanisms: 

\begin{itemize}
    \item \textbf{Missing Completely at Random (MCAR)}: MCAR occurs when the missingness is completely unrelated to any variable and is totally random \cite{little2019statistical}. It is the simplest case, yet this strict assumption is rarely satisfied in real-world datasets \cite{raghunathan2004we}.
    \item \textbf{Missing at Random (MAR)}: MAR occurs when the missingness systematically depends on variables with complete information \cite{little2019statistical}.
    \item \textbf{Missing Not at Random (MNAR)}: MNAR occurs when the missingness depends on the variable of interest itself or on other unknown variables \cite{little2019statistical}.
\end{itemize}

Different missing data mechanisms have different practical implications. These are discussed in conjunction with different missing data treatment methods in the following section. This background section concludes with a discussion of the role of subjectivity in missing data mitigation.

\subsection{Missing data treatment}

There are many available missing data treatment methods \cite{emmanuel2021survey, kang2013prevention, pham2024missing}. This subsection reviews some common methods, their advantages and drawbacks:

\begin{itemize}
    \item \textbf{Complete case analysis (CCA)}: CCA means discarding all records with missing data and only training models on complete data \cite{white2010bias}. It is also known as available case analysis or listwise deletion. CCA is easy to use as it is the default method in many statistical packages \cite{white2010bias}. However, CCA is criticized for two major drawbacks. First, it could discard a significant amount of data, so it is sometimes considered inefficient \cite{white2010bias}. Second, if the data is not MCAR, CCA is not a random subsample of the original sample and may lead to biases in some cases \cite{little2024comparison}. For example, if in a hypothetical medical dataset, women are more likely to have missing data than men, applying CCA to the dataset means discarding most records that belong to women, potentially making model prediction results less accurate for women.

    \item \textbf{Mean imputation}: Mean imputation fills in missing values with the average value of the feature. It is the easiest and one of the most widely used imputation methods \cite{jadhav2019comparison}. However, if a feature contains many missing data points, mean imputation could lead to changes in data distribution. For example, the standard deviation could become smaller after imputation \cite{jadhav2019comparison}. This change can disturb the relations between variables and induce bias, so mean imputation should not be used when data is not MCAR \cite{bennett2001can, van2018flexible}. Similarly, filling in missing values with median, mode, or an arbitrary value (e.g., 0) can also introduce significant bias when data is not MCAR.

    \item \textbf{Last observation carried forward (LOCF)}: This method replaces the missing values with the last observed values. This is a common approach for treating longitudinal data and time series data. However, researchers found that this method introduces significant bias when data is not MCAR \cite{lachin2016fallacies}.
    
    \item \textbf{Model-based imputation}: Recognizing the limitations of mean imputation, some researchers proposed to train models (e.g., regression models, k-nearest neighbors, neural networks, and so on) based on available data to predict missing values \cite{joel2022review}. Model-based imputation methods can improve performance in some cases, but these methods make assumptions regarding relationships among the features \cite{joel2022review}. For example, imputing missing values using regression models assumes a linear relationship between the features \cite{joel2022review}. Regression imputation will introduce biases into the dataset if the assumption is not true \cite{joel2022review}. 
    
    More generally, all model-based imputation methods assume MAR. By definition, MAR means the missingness systematically depends on variables with complete information, which is why models trained on variables with complete information can be used to predict missing values. However, if the data is not MAR, model-based imputation will introduce biases.

    \item \textbf{Multiple imputation}: Taking the uncertainty of missing data into consideration, many statisticians advocate for multiple imputation \cite{kenward2007multiple, rubin2004multiple}. Multiple imputation imputes missing values multiple times, producing multiple complete datasets with imputed values. Each imputed dataset is analyzed separately, and the results are pooled together using statistical rules \cite{jadhav2019comparison, rubin2004multiple}.

    While multiple imputation has long been established in the statistics community, it has only recently gained attention in the ML community. Ensemble learning is one way of incorporating multiple imputation into the ML training process \cite{aleryani2020multiple}. One ML model is trained on each dataset produced by multiple imputation \cite{aleryani2020multiple}. Then, the models are grouped as an ensemble \cite{aleryani2020multiple}. While research showed that ensemble learning based on multiple imputation can improve classification accuracy \cite{aleryani2020multiple}, in practice, this method can be computationally intensive, and it reduces the model's interpretability.  Another way is to use multiple imputation by chained equations (MICE) \cite{white2011multiple}. However, MICE assumes MAR and is sensitive to departures from the MAR assumption \cite{white2011multiple}.

    \item \textbf{Missing-indicator method}: This method is used in conjunction with the imputation methods. It adds binary dummy variables to indicate which values were originally missing and have been imputed \cite{van2023missing}. This method is easy to implement and can be particularly useful in cases of informative missingness, which often occurs when data is MNAR \cite{van2023missing}. For instance, in a hypothetical financial dataset, if people with lower credit scores are less likely to report their credit scores, then whether a person's credit score is missing is informative. 
    
    However, if many variables have missing data, the missing-indicator method adds more variables, increasing the dimensionality of the dataset \cite{van2023missing}. High-dimensional datasets are more likely to induce overfitting and reduce the interpretability of the ML models trained.

    \item \textbf{ML algorithms that can take input with missing data}: Several types of ML algorithms, such as generalized additive models and decision trees, can automatically handle missing data \cite{mctavish2024interpretable}. Decision trees and other tree-based algorithms, such as XGBoost, can handle missing data in several different ways. First, when splitting on features that involve missing data, the tree-based algorithms evaluate whether sending all missing values to the left node or to the right node leads to better performance. This method is used by scikit-learn, one of the most popular Python ML libraries, when the parameter ``splitter" is set to ``best" for the DecisionTreeClassifier or the  DecisionTreeRegressor class \cite{JMLR:v12:pedregosa11a}. XGBoost also uses this method by default \cite{chen2016xgboost}.
    
    Alternatively, when ``splitter" is set to ``random," scikit-learn's ExtraTreeClassifier and ExtraTreeRegressor classes employ the following strategy \cite{JMLR:v12:pedregosa11a}:
    \begin{quote}
        When splitting a node, a random threshold will be chosen to split the non-missing values on. Then the non-missing values will be sent to the left and right child based on the randomly selected threshold, while the missing values will also be randomly sent to the left or right child. This is repeated for every feature considered at each split. The best split among these is chosen.
    \end{quote}

 Depending on the libraries used and the parameters set, tree-based algorithms employ a wide range of missing data treatment methods. For example, in XGBoost, when ``booster" is set to ``gblinear," all missing values are treated as zeros \cite{chen2016xgboost}. This is equivalent to fixed-value imputation, which could introduce significant bias when data is not MCAR. Therefore, if the underlying missing data treatment mechanisms are not carefully reviewed, relying on existing implementations of ML algorithms to automatically handle missing data can still introduce bias.
\end{itemize}

\subsection{Subjectivity in missing data treatment}
Since there are many different missing data treatment methods to choose from, researchers debate which method is the best. For instance, there is much debate around whether multiple imputation or CCA is better. Many stand in favor of multiple imputation \cite{ali2011comparison, white2010bias}, although others found that multiple imputation is not always preferable to CCA \cite{mukaka2016using}. Recognizing the complicated situation, some researchers state that the relative gain of multiple imputation over CCA depends on how much information is contained for the analysis of interest in incomplete cases, which can vary dramatically depending on the context \cite{little2024comparison}.

Despite many disagreements, most researchers agree on two points. First, it is important to assess the patterns and possible reasons for missing data since some missing data treatment methods are only valid under certain missing data mechanisms \cite{joel2022review, little2024comparison, nijman2022missing, raghunathan2004we}. Second, mean, median, mode, or fixed-value imputation should not be used unless the data is MCAR \cite{bennett2001can, jadhav2019comparison, van2018flexible}. However, many ML researchers do not follow these suggestions. A study reviewing 152 ML-based clinical prediction model studies found that a substantial amount (56 papers) did not report any information on missing data \cite{nijman2022missing}. Out of the 96 papers that reported on missing data, only seven reported possible reasons for missing data \cite{nijman2022missing}.

It remains unknown how ML researchers and engineers choose missing data treatment methods, what factors could affect their choices, and how familiar they are with statistical theories of missing data. To investigate these, we conducted a survey of 70 ML researchers and engineers. The next section describes our survey method.

\section{Method}

Our survey contained a mixture of free-response questions and multiple-choice questions about missing data treatments, as well as several demographic and logistic questions. On average, the survey took about 12 minutes to complete. The complete survey is in Appendix A. 

After IRB approval, we distributed the survey through university listservs and social media. To avoid potential selection bias, we advertised the survey as a generic survey about practices in ML training and did not mention ``missing data" in our recruitment materials. We targeted participants in both academia and industry, which included graduate students and faculty in computer science and related fields, research scientists and machine learning engineers. Our selection criteria specified that the participants must be at least 18 years old with some prior experience (e.g., taken a class, done a project/internship, etc.) in ML. As an incentive and an appreciation for the participants' time, a raffle randomly selected eight participants, and each was awarded a \$25 Amazon gift card. Participants' answers did not affect their winning chances.

\section{Results}

\begin{figure}
    \centering     \includegraphics[width=0.6\linewidth]{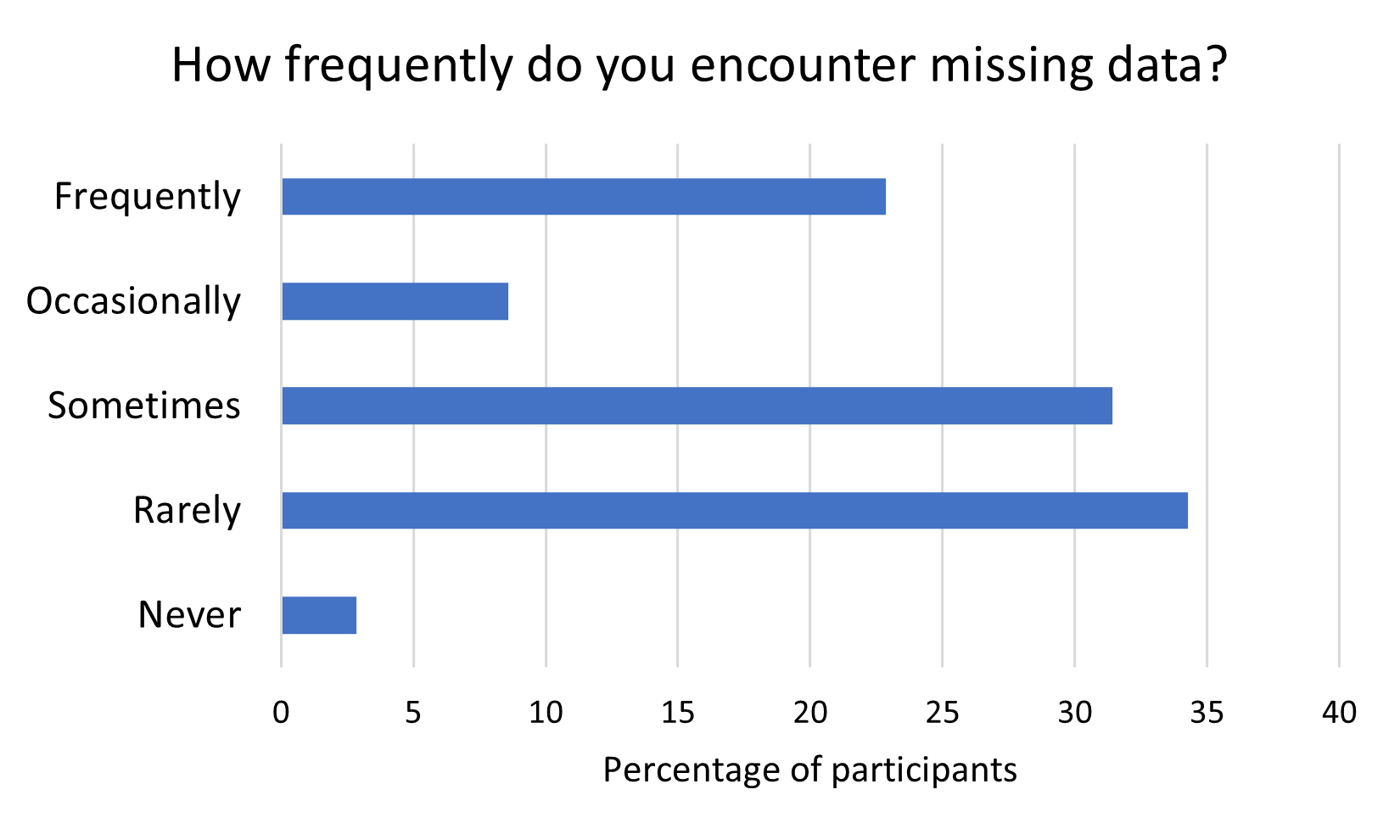}
    \caption{Prevalence of missing data}
    \label{fig:missing-prevalence}
\end{figure}

We collected 70 valid survey responses from 32 graduate students, 20 faculty, and 18 industry professionals with ML-related jobs. The full demographic information is included in Appendix B. The responses showed that missing data is prevalent in both academia and industry. Eighteen participants (25.71\%) have worked with datasets with missing data in internships or full-time jobs in the industry, 29 participants (41.43\%) in academic research, and 30 participants (42.86\%) in coursework. As shown in Figure \ref{fig:missing-prevalence}, only two participants (2.86\%) never encountered missing data.

\begin{figure}
    \centering
    \includegraphics[width=1\linewidth]{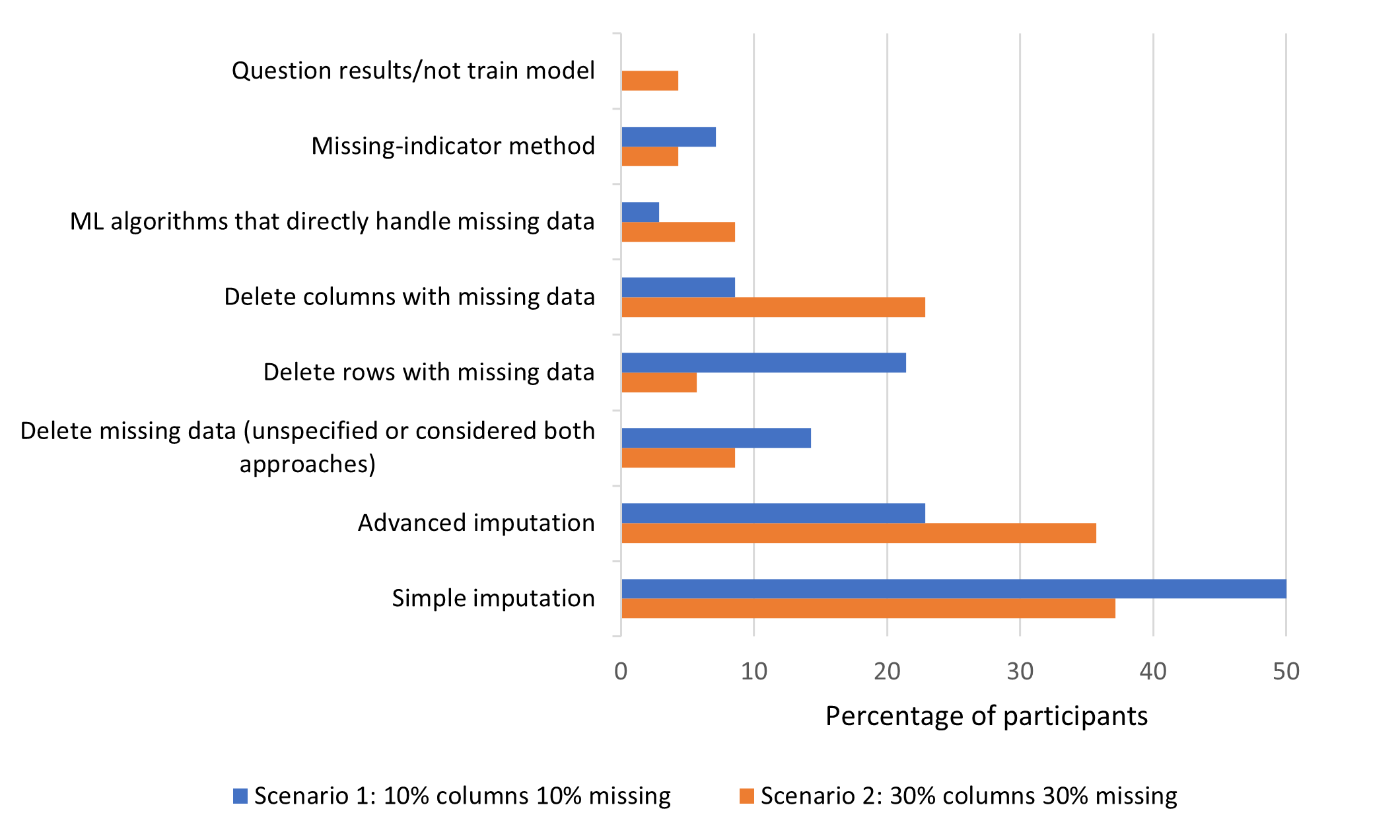}
    \caption{Missing data treatment methods mentioned in free response questions. Some participants mentioned multiple methods, so the percentages do not add up to 100\%.}
    \label{fig:free-response}
\end{figure}

To understand how ML researchers and practitioners treat missing data in different scenarios, we began the survey with two free-response questions:

\begin{enumerate}
    \item Scenario 1: Suppose you are trying to train a machine learning model on tabular data. Your dataset has 100 columns and 10,000 rows. Ten columns have 10\% data missing. How would you address the missing data?
    \item Scenario 2: Now suppose thirty columns have 30\% data missing. All other conditions remain the same. How would you address the missing data?
\end{enumerate}

For each response, we manually classified the missing data treatment methods that the participants mentioned. Mean imputation, median imputation, mode imputation, fixed-value imputation, and last observation carried forward were classified as ``simple imputation." Simple imputation methods are based on relatively simple rules. They are easy to implement but are likely to introduce biases when data is not MCAR. Within the simple imputation category, mean imputation was mentioned most frequently, with 28 mentions in total. Median imputation was mentioned ten times, and all other methods were each mentioned five times or less. 

Model-based imputation and multiple imputation were classified as ``advanced imputation." Advanced imputation methods require more computational resources, and they all share the MAR assumption as they predict missing features based on known features. Participants mentioned a wide variety of ways to predict missing values. K-nearest neighbors (KNN) and regression models were the most popular, with eight mentions each. Multiple imputation was mentioned six times, and the Bayesian method was mentioned twice. Expectation-maximization (EM) algorithm, decision tree, neural network, autoencoder, and ensemble learning were each mentioned once.

Figure \ref{fig:free-response} summarizes the results of the free-response questions. In scenario 1, 50\% of participants considered using simple imputation, making it the most popular method. Advanced imputation was also popular, and 22.86\% considered it. Deleting rows with missing data (21.43\% of participants) was much more popular than deleting columns with missing data (8.57\% of participants). In scenario 2, approximately an equal number of participants considered simple imputation (37.14\% of participants) and advanced imputation (35.71\% of participants). Deleting columns with missing data (22.86\% of participants) was much more popular than deleting rows with missing data (5.71\% of participants). 

\subsection{Similarity between scenarios}

Simple imputation was the most popular method in both scenarios. However, 36 participants (51.4\%) considered using simple imputation without mentioning that they would check the missing mechanism first in at least one scenario. Since simple imputation can introduce significant bias, statisticians have advised not to use it unless the data is MCAR \cite{bennett2001can, jadhav2019comparison, van2018flexible}. Given that real-world datasets rarely satisfy the MCAR assumption, the models produced by these individuals likely have questionable validity if the original datasets contain missing data.

For scenario 2 with significantly more missing data, only three participants indicated that the amount of missing data would negatively affect the quality and reliability of the ML models, concluding that models should not be trained at all or at least should not be used in critical contexts. One participant wrote, ``The data set might be useful to test various data analysis techniques, but I would not use the outcomes for anything wherein catastrophic or hazardous outcomes could occur." However, the vast majority of participants did not consider ``not train model" as an option, even in this extreme case. None mentioned this concern in scenario 1.

\subsection{Differences between scenarios}

The amount of missing data present in the dataset significantly affected the participants' choices. When the scenario involved more missing data, participants were more likely to delete columns rather than rows with missing data. One reason is that some participants thought that columns with 30\% missing data could not be accurately imputed, so they would delete those columns altogether. Another reason is that some considered columns with more missing data unreliable. For example, one participant wrote in scenario 2, ``I'd start considering whether the other 70 columns have sufficient information for the domain at hand; if 30 of the variables are so unreliable, maybe we're better off using the 70 reliable ones." However, missingness does not always correlate with reliability, especially in the case of informative missingness where whether a feature is missing or not entails significant information about the case. Participants who delete columns without checking contexts run the risk of deleting important features.

Another notable difference is that when more missing data were present, participants were more likely to use advanced imputation. Thirteen participants  (18.6\%) would use deletion methods or simple imputation in scenario 1 but would use advanced imputation in scenario 2. One reason is that some participants thought small amounts of missing data would not have much impact on model training, so in those cases, they would use methods that are easier to implement, such as deletion and simple imputation. However, when more missing data were present and could have a stronger effect, they would make more effort to treat missing data. For instance, in scenario 1, one participant wrote, ``I would probably just leave out those rows to train the model. Easy." In scenario 2, the same participant wrote, ``Okay now this is an issue. I'd probably train on the data I have, then use that model to fill in my missing data, then train on the complete model." The phrase ``now this is an issue" suggested that missing data was not considered an issue in the first scenario. However, not all participants took scenario 2 more seriously. Another participant who also chose to delete rows in scenario 1 employed the same strategy in scenario 2, saying, ``I think we can still get away with dropping the rows with missing data."

Concerns about data loss also affected some participants' choices. When more missing data were present, deletion methods would delete a large amount of data. Sixteen participants (22.86\%) would use deletion methods in scenario 1 but not in scenario 2. For instance, one participant wrote, ``Since a larger portion of columns has missing data, I'd be more cautious about deleting rows. Only remove those rows where the missing values don't seem to carry significant information, to avoid severely shrinking the dataset." However, not all participants shared this concern about data loss. In scenario 2, another participant wrote, ``The data is already overfitting."

\subsection{Familiarity with theories and methods}

\begin{figure}
    \centering    \includegraphics[width=0.9\linewidth]{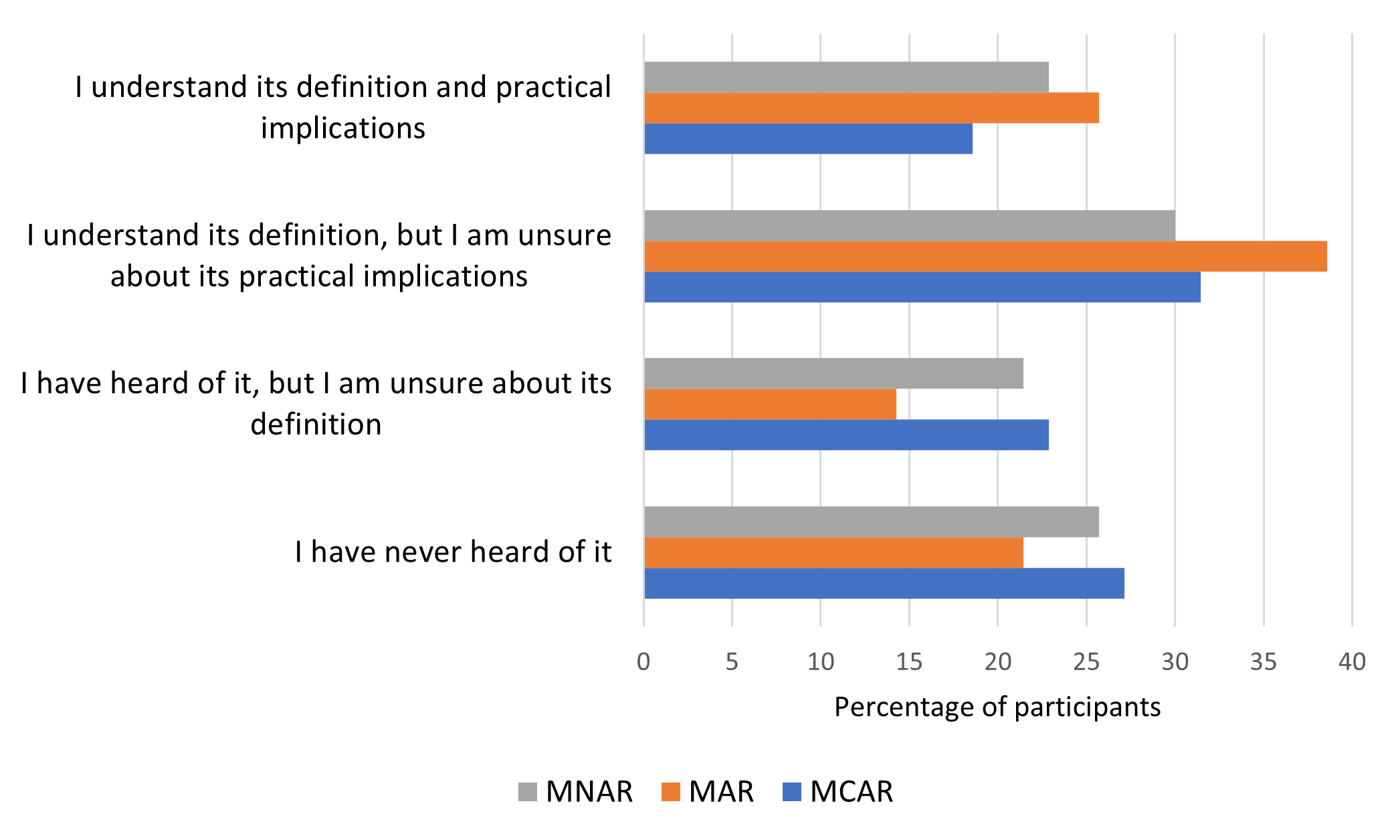}
    \caption{Familiarity with missing data mechanisms}
    \label{fig:familiarity-missing-mechanism}
\end{figure}

\begin{figure}
    \centering    \includegraphics[width=0.9\linewidth]{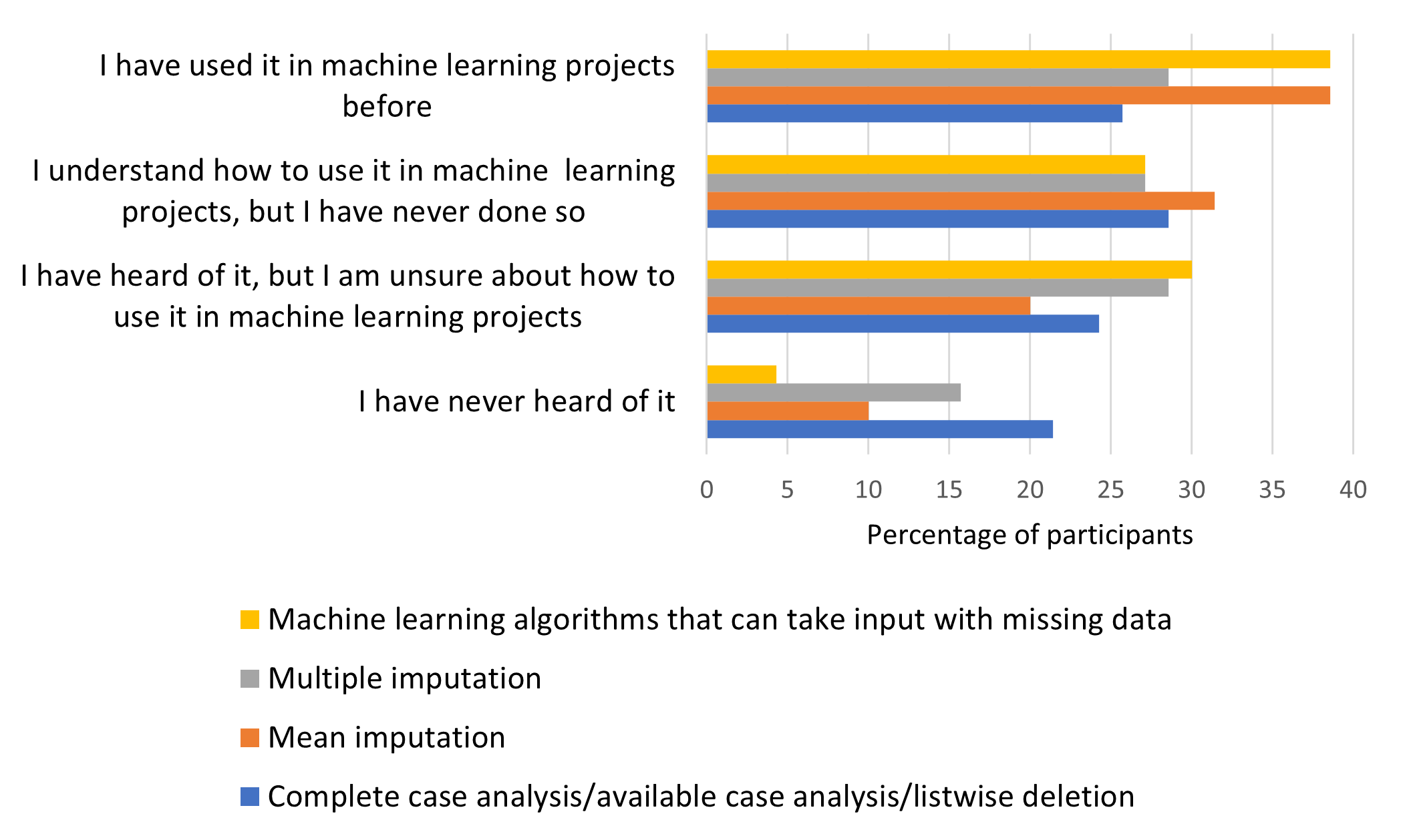}
    \caption{Familiarity with missing data treatment methods}
    \label{fig:familiarity-treatments}
\end{figure}

The survey asked the participants how familiar they were with each of the missing data mechanisms: MCAR, MAR, and MNAR. Figure \ref{fig:familiarity-missing-mechanism} summarizes these results, which shows that the majority of participants do not fully understand missing data mechanisms. Fourteen participants (20\%) have never heard of any missing data mechanisms, and only six participants (8.57\%)  indicated that they understood the definitions and practical implications of all three missing data mechanisms. However, out of these six participants, only one mentioned checking missing data mechanisms in the free-response questions. This suggests that the self-reported familiarity with missing data mechanisms does not correlate with participants' choices in practical situations. The vast majority (80\%) of participants did not mention investigating reasons and patterns of missing data, even if some self-reported that they fully understood the missing data mechanisms.

The survey also asked the participants how familiar they were with different missing data treatment methods. Figure \ref{fig:familiarity-treatments} summarizes the results. Participants were most familiar with mean imputation and ML algorithms that can take input with missing data. However, while 27  participants (38.57\%) indicated that they have previously used ML algorithms that can take input with missing data, only six  participants (8.57\%) mentioned this method in at least one free response questions. It is noteworthy that while many claimed to be familiar with this method, very few demonstrated this knowledge in practice.

\section{Discussion}

The results showed that the participants generally lacked familiarity with the definitions and practical implications of the missing data mechanisms. Consequently, most participants could not make informed decisions regarding missing data treatment, potentially introducing biases into the datasets and ML models. Moreover, we did not find any correlations between the participants' responses and their backgrounds, suggesting that this is a prevalent problem for ML modelers with all backgrounds. More years of experience in ML did not help participants make more informed decisions. 

Therefore, better education about missing data is needed to mitigate this situation. Although missing data is prevalent in both academia and industry, over half (57.14\%) of the participants did not gain experience treating missing data through coursework. We suggest that missing data treatment be an integral part of the ML curriculum. Educators should focus on teaching the practical implications of the missing data mechanisms. For each missing data mechanism, 30\% or more participants indicated that they understood the definitions but were unsure about the practical implications. Although researchers in the field of missing data methodology have provided several practical suggestions \cite{emmanuel2021survey, kang2013prevention, pham2024missing}, these suggestions were not well-known to our participants.

Moreover, the ML community needs better data and model reporting standards. Our survey results revealed that missing data treatment is very subjective. Given the same scenario description, the participants considered a wide variety of approaches. Furthermore, participants could not agree on how much missing data would be too much and would require a different approach. The subjectivity threatens the repeatability and reproducibility of ML models. This problem is especially important when ML is applied to make scientific discoveries, as the scientific community cannot assess the validity of research based on ML models. Therefore, ML-based research should be required to report the amount of missing data in training and testing sets, possible reasons for missing data, missing data mechanisms, and missing data treatment methods used. 

Lastly, it will be beneficial to have better missing data analysis tools. Tools that automatically analyze missing data patterns and make recommendations have the potential to help ML modelers make more informed decisions. Currently, most statistical tools use CCA as the default method to handle missing data \cite{white2010bias} and do not make recommendations on alternative options.

One limitation is that our study relied on self-reported survey responses. However, the participants' choices in practical situations might be different from their self-reported answers. Further studies are needed to investigate ML modelers' preferences in real-world situations.

\section{Conclusion}

Our survey showed that missing data treatment is a very subjective process. Factors that influenced the participants' choices include the amount of missing data present and concerns about data loss. Our survey also revealed that most participants could not make informed decisions due to a lack of familiarity with missing data mechanisms. Over half of the participants made choices that could induce biases in the datasets and ML models. To enhance the validity of ML models, we advocate for better education on missing data, more standardized missing data reporting, and better missing data analysis tools.

\bibliography{bib}

\appendix

\section{Survey}

\includegraphics[width=0.9\linewidth]{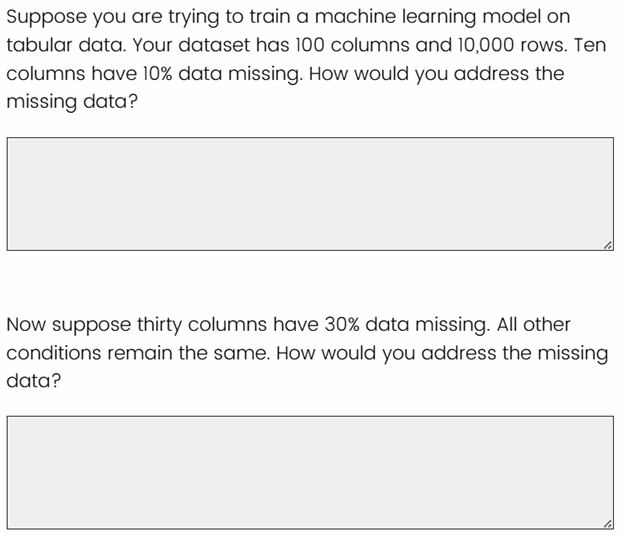}

\vspace{1 cm}

\includegraphics[width=0.9\linewidth]{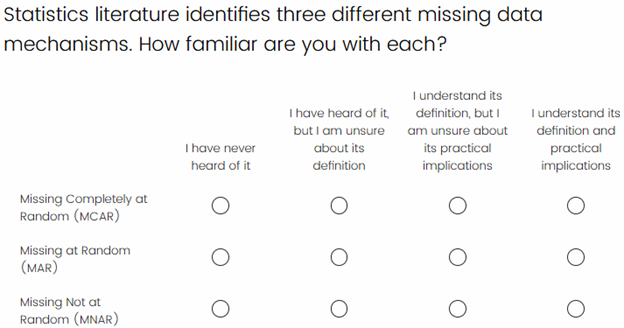}

\vspace{1 cm}

\includegraphics[width=0.9\linewidth]{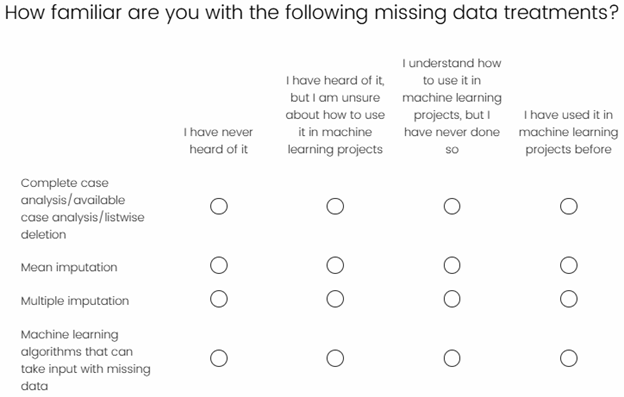}

\vspace{1 cm}

\includegraphics[width=0.9\linewidth]{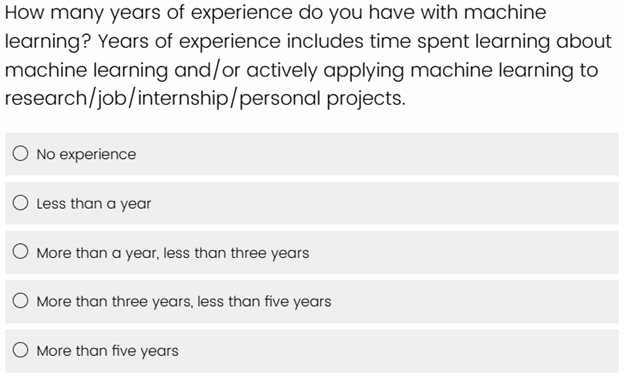}

\vspace{1 cm}

\includegraphics[width=0.9\linewidth]{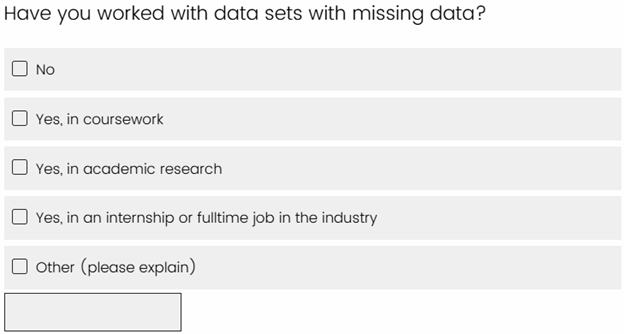}

\vspace{1 cm}

\includegraphics[width=0.9\linewidth]{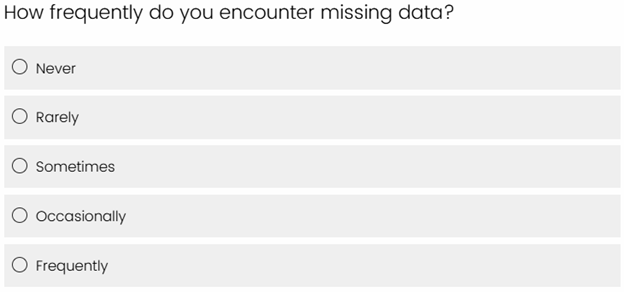}

\vspace{1 cm}

\includegraphics[width=0.9\linewidth]{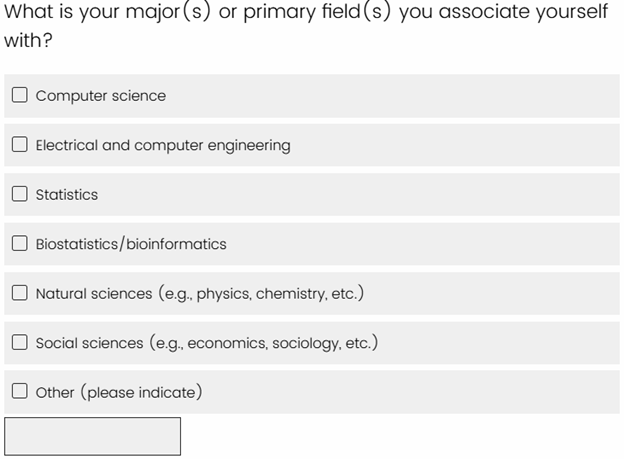}

\vspace{1 cm}

\includegraphics[width=0.9\linewidth]{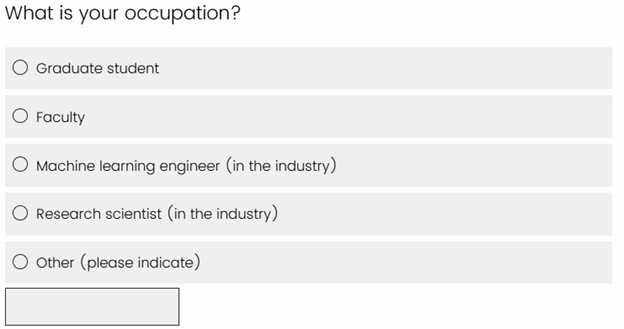}

\vspace{1 cm}

\includegraphics[width=0.9\linewidth]{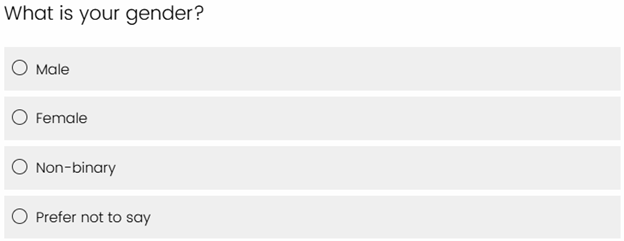}

\vspace{1 cm}

\includegraphics[width=0.9\linewidth]{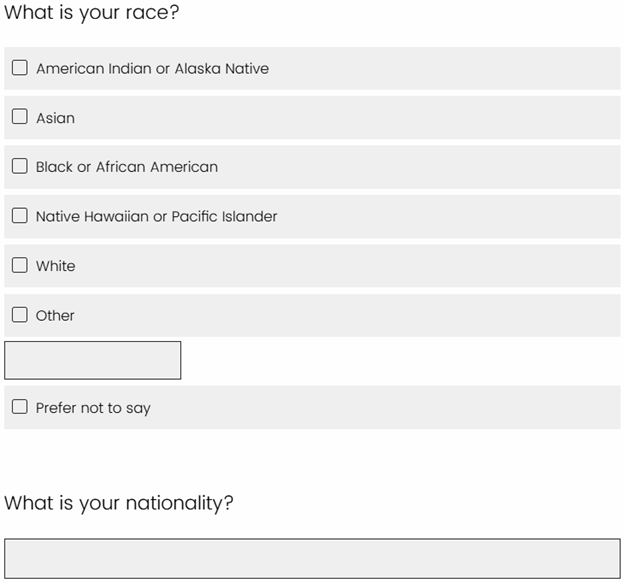}

\section{Demographics}

\begin{table}[htp]
    \centering
    \begin{tabular}{lc}
        \toprule
        Years of ML experience & Number of participants \\
        \midrule
        Less than a year &  13\\
        More than a year, less than three years & 22\\
        More than three years, less than five years & 21\\
        More than five years & 14\\
        \bottomrule
    \end{tabular}
\end{table}

\begin{table}[htp]
    \centering
    \begin{tabular}{lc}
        \toprule
        Occupation & Number of participants \\
        \midrule
        Graduate student &  32\\
        Faculty & 20\\
        Machine learning engineer (in the industry) & 8\\
        Research scientist (in the industry) & 8\\
        Professionals with other titles in the industry & 2\\
        \bottomrule
    \end{tabular}
\end{table}

\begin{table}[htp]
    \centering
    \begin{tabular}{lc}
        \toprule
        Field(s) & Number of participants \\
        \midrule
        Computer science &  43\\
        Electrical and computer engineering & 11\\
        Statistics & 6\\
        Biostatistics/bioinformatics & 7\\
        Natural sciences (e.g., physics, chemistry, etc.) & 2\\
        Social sciences (e.g., economics, sociology, etc.) & 6\\
        Other & 4\\
        \bottomrule
    \end{tabular}
\end{table}

\begin{table}[htp]
    \centering
    \begin{tabular}{lc}
        \toprule
        Gender & Number of participants \\
        \midrule
        Male &  45\\
        Female & 18\\
        Non-binary & 0\\
        Prefer not to say & 7\\
        \bottomrule
    \end{tabular}
\end{table}

\begin{table}[htp]
    \centering
    \begin{tabular}{lc}
        \toprule
        Race(s) & Number of participants \\
        \midrule
        American Indian or Alaska Native &  8\\
        Asian & 20\\
        Black or African American & 9\\
        Native Hawaiian or Pacific Islander & 1\\
        White & 26\\
        Other & 1\\
        Prefer not to say & 5\\
        \bottomrule
    \end{tabular}
\end{table}

\begin{table}[htp]
    \centering
    \begin{tabular}{lc}
        \toprule
        Nationality & Number of participants \\
        \midrule
        United States &  40\\
        Asian countries & 16\\
        African countries & 3\\
        European countries & 2\\
        Prefer not to say & 9\\
        \bottomrule
    \end{tabular}
\end{table}

\end{document}